\documentclass[13pt]{article}
\usepackage{geometry}
\usepackage{amsmath,amssymb,amsthm}
\usepackage{physics}
\usepackage{scrextend}
\usepackage{sectsty}
\usepackage{graphicx}
\usepackage{fancyhdr}
\usepackage{lipsum}
\usepackage{xcolor}
\usepackage{pgfplots}

\usepackage{enumitem}

\newcommand{\inv}{^{-1}}

\providecommand{\bf}[1]{\mathbf{#1}}

\usepackage{thmtools}
\usepackage{hyperref}
\usepackage{cleveref}
\usepackage{thmtools}
\usepackage{mathrsfs}
\newcommand{\sB}{\mathscr{B}}
\newcommand{\sG}{\mathscr{G}}
\declaretheorem[numberwithin=section]{theorem}
\newtheorem*{remark}{Remark}
\newtheorem*{setup}{Setup}

\newtheorem*{application}{Application}

\newtheorem*{claim}{Claim}

\title{Applications of the Theory of Aggregated Markov Processes in Stochastic Learning Theory}
\author{Fangyuan Lin\\
University of California, Berkeley\\}
\date{Year: 2023}

\begin{document}

\maketitle

\begin{abstract}
A stochastic process that arises by composing a function with a Markov process is called an aggregated Markov process (AMP). The purpose of composing a Markov process with a function can be a reduction of dimensions, e.g., a projection onto certain coordinates. The theory around AMP has been extensively studied e.g. by Dynkin, Cameron, Rogers and Pitman, and Kelly, all of whom provided sufficient conditions for an AMP to remain Markov. In another direction, Larget provided a canonical representation for AMP, which can be used to verify the equivalence of two AMPs. \\

The purpose of this paper is to describe how the theory of AMP can be applied to stochastic learning theory as they learn a particular task.
\end{abstract}

\section{Learning Theory Background}
Stochastic learning theory serves to provide a stochastic model of the process of behavior modification in animals or people. Learning has a stochastic nature (at least it appears to), e.g., a person asked to memorize a list of words sometimes recall more words in an early trial than in a later trial. 

\begin{quote} Even in simple repetitive experiments, the sequences of choices made by the subject are typically erratic, suggesting a probabilistic nature of learning. (Sternberg 1963) \end{quote}

The models of stochastic learning theory serve to describe the trial-to-trial changes in probabilities to make certain responses and are usually either Markov or a function of a Markov process. \\
\begin{itemize}
    \item AMPs are generally not Markov, so the aggregated model does not necessarily enjoy Markovanity.
    \item The aggregating functions are assumed to be onto. Also, we are only interested in many-to-one aggregating functions since a process derived by a one-to-one aggregating function always contains the same information as the underlying process that is relevant to the next state of the derived process. 
\end{itemize}

\subsection{Stochastic Learning Theory Framework}

The following is a brief description of a framework (Norman 1968) for the construction of a learning model.\\
\begin{itemize}
\item 
Consider the process $X_n$ with state space $(X, \sB)$. Here $X_n$ represents the subject's response probability on trial $n$.
\item
Consider also the event space $(E,\sG)$, which represents the set of possible outcomes of a trial that can happen to the subject. 
\item 
The process $E_n$ represents the outcome on trial $n$ and usually contains which response the subject makes as one of its coordinates.
\item
Let $\mu: X\times E\to X$ be a measurable function with respect to $\sB\times\sG$ and $\sB$. $\mu$ has the role of transforming $X_n$ based on $E_n$, i.e., $X_{n+1} = \mu(X_n, E_n)$. In other words, the trial-to-trial change of response-probability is governed by the deterministic operator $\mu$, which transforms the response-probability depending on the event that occurs in the previous trial. (Most models are assumed to be time-homogeneous, meaning $\mu$ does not depend on the trial $n$.)
\item
Let $P$ be a stochastic kernel on $X\times\sG$ and $P$ has the role that $P(X_n,A) = P(E_n\in A|X_n,E_{n-1,...}$ for $A\in\sG$.
\end{itemize}

The system $((X,\sB), (E,\sG),P,\mu)$ is called a {\em random system with complete connection}, and it is a formalization of learning model.\\

\section{Applications of the Theory of Aggregated Markov Processes}

\begin{setup}Let $Z_n$ be a Markov chain on the
countable state space $S_Z$ with transition probability function $P$. Let 
$Y_n = f(Z_n)$, for $f: S_Z \to S_Y$ be the aggregated process.
\end{setup}

\begin{theorem}[Burke and Rosenblatt (1958) Simplest condition to preserve Markovianity]\label{simplest} Suppose that
\begin{align}
    P\{Y_{n+1}|Z_{n} = z_n\} = P\{Y_{n+1}|Y_n=f(z_n)\} = P\{Y_{n+1}|Z_n\in f^{-1}(f(z_n))\},
\end{align}
then $Y$ is Markov.
\end{theorem}
\begin{remark} The condition of the theorem says that $Y_n$ contains the same information as $Z_n$ does for predicting $Y_{n+1}$.
\end{remark}
\begin{remark} The Markov process $Y$ is not necessarily homogeneous.
\end{remark}

Here is a straight-forward application of the above theorem.

\begin{application}[Theorem 2.1]
     {\em A model based on the continuous pattern model of Suppes (1959).} A subject is asked to predict the location on the rim of a disk that will be illuminated. Let $Y_n$ be the subject's prediction and let $Z_n$ be the actual spot of light that will be revealed after the subject makes their choice. Assume the set of locations $Y$ is discrete so that the theorem applies (we can always consider a higher granularity for better accuracy).
     
     Let $S$ be a set of $N$ stimulus patterns. The state of learning $x: S\to Y$ is a function mapping a stimulus pattern to a point on the rim of the disk, and $Y_n$ has distribution $\lambda_s(x_s,dy)$.
     
     On each trial, the subject samples a stimulus pattern $s$ uniformly randomly from $S$ (i.e. with probability $\frac{1}{N}$) and makes a prediction according to $\lambda_s(x_s,dy)$.
     
     Given a prediction $y$, the outcome $z$ has distribution $\pi(y,dz)$. In this model, the event process is $(S_n, Y_n, Z_n).$
     
    After observing the spot illuminated $z$, $x_s$ is updated to $z$. The transformation $u$ plays the role of updating $x_s$: \begin{align*}
        u(x_s,e) &= z \\
        u(x_t,e) &= x_t &\text{for $t\neq s$}
    \end{align*}

    Note that the original model also considers the probability of successful conditioning and I neglect this aspect for simplicity.
    
    One observation is that the transformation of $x$, the state of learning, does not depend on the prediction $y$ that the subject just made, so we have the motivation to reduce the model from $(x,s,y,z)$ to $(x,s,z)$, i.e. we do not keep track of the subject's response.
    
    To see the reduced model still has the Markov property, we can apply Theorem 2.1. The distribution of $S_n$ is always the uniform distribution on $S$. The next state of learning $X_n$ depends only on $S_n$ and $Z_n$ and is independent of $Y_n$, as noted previously. The distribution of $Z_{n+1}$ does depend on $Y_{n+1}$; however, the distribution of $Y_{n+1}$ follows $\lambda(X_{n+1},dy)$ and $X_{n+1}$ does not depend on $Y_n$.
    
    Therefore, the reduced model is Markov by Theorem 2.1.
\end{application}

Next, we introduce a theorem applicable to process with uncountable state space. 
\subsection{Theorem 2.2}
\begin{theorem}[Cameron (1973)]
    If for $z, z'\in S_Z$, if $f(z)=f(z')$, then for any $n$ and event $B_n\subseteq S_Y$, the $n$-step transition probability $P_n(z, f\inv(B_n)) = P_n(z', f\inv(B_n))$,\\
    then $Y_n$ is Markov. 
\end{theorem}

\begin{remark}
    The condition says that what matters is where the starting point is mapped to, not the starting point itself.
\end{remark}
\begin{application}[Theorem 2.2]
    {\em Reduction of dimensionality in the  Zeaman-House-Lovejoy (ZHL) Model.}
    Suppose we have a T-maze where on each trial, one of the left and right arms is chosen to be bright (white) uniformly at random. 
    
    A rat then chooses which arm to go to. On each trial, we assume that the rat can only pay attention to either brightness or the position of the arm. In this model, the food (reward) is always provided on the black arm. Therefore, the position of the arms is actually completely irrelevant, and the rat needs to learn to pay attention to brightness instead of the position of the arms.\\
    
    We can formulate the stochastic process as follows:\\
    
    Let $\mathrm{br}$ be the event that the rat pays attention to brightness, and $\mathrm{pos}$ be the event that the rat pays attention to the position of the arm. Let $W$ be the event that the rat chooses the white (bright) arm, and let $B$ be the event that the rat chooses the black (dark) arm. \\
    We are interested in:\begin{align*}
        V_n = Pr(\mathrm{br}) &\text{ on trial $n$}\\
        Y_n = Pr(B|\mathrm{br}) &\text{ on trial $n$}\\
        Z_n = Pr(L|\mathrm{pos}) &\text{ on trial $n$}
    \end{align*}
    We now describe the trial-to-trial change of probability: The probability $V$ of attending to brightness is assumed to increase linearly with some parameter if the rat attends to brightness and is fed or does not attend to brightness and is not fed. Otherwise, it decreases. For example, \\
    $V_{n+1} = u(V_n, e) = \begin{cases}
        V_n + \phi_1 (1-V_n) & e = (B,W)\mathrm{br}B\\
        V_n - \phi_2 V_n & e = (B,W)\mathrm{br}W \\
        \text{and so on}
    \end{cases}$\\
    where $\phi_1, \phi_2 \in (0,1)$ are parameters representing how fast the rat learns from the presence of reward.\\
    
    The conditional response probabilities $Y$ and $Z$ are also changed according to the rat's attention:
    For example, if the rat pays attention to brightness, the $Y$ should always be increased (linearly in this model) since we assume that the rat learns from the presence or lack of reward! In this case, $Z_n$ does not change at all, which is intuitive. If the rat pays attention to position, $Z_n$ is increased (linearly in this model) only if the left arm is, luckily, the black arm. In this case, $Y_n$ does not change at all, which is intuitive.\\
    
    The stochastic process that we have an interest in is $X_n = (V_n, Y_n, Z_n)$, which represents the response probabilities and it has a continuous state space, so Theorem 2.1 does not apply. The event process is $E_n=(S_n, A_n, R_n)$ where $S_n$ represents the location of the black arm, $A_n$ is either $\mathrm{br}$ or $\mathrm{pos}$, representing the rat's attention, and $R_n$ is $B$ or $W$, representing which arm the rat chooses on trial $n$.
    
    \begin{claim}
        As the assumption of the model says, the location of the arms is irrelevant. We can ask whether we can reduce the dimensionality of the model to reduce its complexity while preserving the Markovianity. The answer is yes. Theorem 2.2 applies in this example.
    \end{claim}
\end{application}

\begin{proof}
    Consider the coordinate projection $\Phi(v,y,z)=(v,y)$. We apply $\Phi$ to $X_n$ so that from now on, we stop caring about the probability of the rat choosing the left arm when it pays attention to the position of the arms, motivated by the fact we just want to know how well the rat learns to pay attention to brightness and go to the black arm.\\
    
    Note that we can in fact also apply the projection $\Psi(s,a,r)=(a,r)$ to $E_n$ so that we stop keeping track of where the black arm is. After all, it is the brightness, not the location, that matters.\\
    
    To apply Theorem 2.2, we need to show that for any $n$ and $A_n\subset[0,1]\times[0,1]$ measurable,\\
    $P_n((v,y,z), f\inv(A_n)) = P_n((v,y,z'), f\inv(A_n))$, for any $z$ and $z'$, since $\Phi(v,y,z)=\Phi(v,y,z')=(v,y)$ for any $z$ and $z'$.\\
    
    Recall that $X_{n+1}$ is determined by a transformation that takes $E_n$ and $X_n$ (i.e., $X_{n+1}=u(X_n, E_n)$).
    \begin{itemize}
    \item If $\mathrm{b}$, $V_n$ is changed according to the occurrence of $B$: it increases linearly if the rat receives the reward by choosing to go to the black arm (which happens with probability $Y_n$) and decreases linearly with probability $1-Y_n$.
    \item If $\mathrm{pos}$, $V_n$ is increased linearly if the rat does not receive the reward, which happens with probability: \begin{align*}
        Pr(W_n|X_n,pos)&=Pr(L|X_n,\mathrm{pos},(W,B))Pr((W,B))+Pr(R|X_n,\mathrm{pos},(B,W))Pr((B,W)) \\
        &\text{by conditioning on the position of the white and black arms}\\
        &\text{(where $(W,B)$ stands for the case where the white arm is on the left)}\\
        &= Z_n\cdot\frac{1}{2}+(1-Z_n)\cdot\frac{1}{2}\\
        &= 1/2
    \end{align*}
    Similarly, $V_n$ is decreased linearly if the rat does receive the reward while paying attention to the position of the arms, which happens with probability $Pr(B_n|X_n,\mathrm{pos})=\frac{1}{2}$.
\end{itemize}
In both cases, $\mathrm{br}$ and $\mathrm{pos}$, the transformation of $V_n$ does not depend on $Z_n$. Also, the occurrence of $\mathrm{br}$ and $\mathrm{pos}$ is dictated by $V_n$ only.\\
\begin{itemize}
    \item If $\mathrm{br}$, the conditional probability $Y_n$ is always increased linearly.
    \item If $\mathrm{pos},$ $Y_n$ remains the same.
\end{itemize}
The transformation of $Y_n$ depends solely on the occurrence of $\mathrm{br}$, which, again, does not depend on $Z_n$.\\
Therefore, $P_n((v,y,z), f\inv(A_n)) = P_n((v,y,z'), f\inv(A_n))$ since as long as we start with the same $v,y$, the distribution of the future of $V_n,Y_n$ is independent of $z,$. We do not need to consider the distribution of the future of $Z_n$, since $f\inv(A_n)=A_n\times[0,1]$, i.e., there is no restriction on what value the future of $Z_n$ takes. The condition of Theorem 2.2 is thus met, and the proof is complete.
\end{proof}

\bigskip
\noindent
{\bf Acknowledgment:} The author thanks Steven N. Evans for helpful discussions during the preparation of this paper.

\newpage
\section*{Reference}
\begin{enumerate}
    \item Bush, R. R., \& Mosteller, F. (1955). Stochastic models for learning. John Wiley \& Sons Inc. https://doi.org/10.1037/14496-000
    \item Cameron, M. A. (1973). A Note on Functions of Markov Processes with an Application to a Sequence of $\chi^2$ Statistics. Journal of Applied Probability, 10(4), 895–900. https://doi.org/10.2307/3212394
    \item Kelly, F. P. (1982). Markovian Functions of a Markov Chain. Sankhyā: The Indian Journal of Statistics, Series A (1961-2002), 44(3), 372–379. http://www.jstor.org/stable/25050326
    \item Norman, M. F. (1970). [Review of Random Processes and Learning., by M. Iosifescu \& R. Theodorescu]. The Annals of Mathematical Statistics, 41(4), 1381–1383. http://www.jstor.org/stable/2240182
    \item Rogers, L. C. G., \& Pitman, J. W. (1981). Markov Functions. The Annals of Probability, 9(4), 573–582. http://www.jstor.org/stable/2243410
    \item Rosenblatt, M. (1971). Markov processes structure and asymptotic behavior. Springer. 
    \item Sternberg, S. (1963). Stochastic learning theory. John Wiley \& Sons Inc.
    \item Suppes, P. (1959). A linear model for a continuum of responses. Stanford University Press. 
\end{enumerate}
\end{document}